\documentclass[letterpaper, 10 pt, conference]{ieeeconf}  

\IEEEoverridecommandlockouts                              

\usepackage[l2tabu,orthodox]{nag}

\usepackage[
backend=biber, 
bibencoding=utf8,
style=ieee, 
sorting=none, 
natbib=true, 
doi=false, 
isbn=false, 
url=false, 
eprint=false, 
maxcitenames=1, 
mincitenames=1,
maxbibnames=5,
]{biblatex}
\usepackage[pdftex,colorlinks]{hyperref}

\usepackage[printonlyused]{acronym}
\acrodef{GNSS}{Global Navigation Satellite System}
\acrodef{RTK}{Real Time Kinematic}
\acrodef{ECU}{Electronic Control Unit}
\acrodef{MPC}{Model Predictive Control}
\acrodef{PID}{Proportional-Integral-Derivative}
\acrodef{IMU}{Inertial Motion Unit}
\acrodef{ANR}{French National Research Agency}
\acrodef{TIARA}{Toward Intelligent Adaptable Robots for Agriculture}
\acrodef{TSCF}{Technologies et systèmes d’information pour les agrosystèmes-Clermont-Ferrand}

\usepackage{siunitx}
\usepackage[all]{nowidow}

\usepackage[dvipsnames]{xcolor}

\usepackage{lipsum}

\usepackage[pdftex]{graphicx}
\usepackage[compatibility=false]{caption}
\usepackage{subcaption}

\usepackage{epstopdf}

\usepackage{import}

\graphicspath{{./latexGoodPractices/}}

\usepackage[dvipsnames]{xcolor}
\usepackage{tikz}
\usetikzlibrary{arrows, arrows.meta, calc, quotes,angles, patterns, intersections}

\usepackage{booktabs}

\usepackage{tabu}
\usepackage{soul}

\usepackage{amssymb,amsfonts,amsmath,amscd}

\usepackage{bm} 

\newcommand{\bbm}{\begin{bmatrix}}
\newcommand{\ebm}{\end{bmatrix}}

\usepackage{xspace}

\title{\LARGE \bfseries
Energy Prediction on Sloping Ground for Quadruped Robots
}

\author{Mohamed Ounally$^{1}$, Cyrille Pierre$^{1}$ and Johann Laconte$^{1}$
\thanks{$^{1}$Université Clermont Auvergne, INRAE, UR TSCF, 63000, Clermont-Ferrand, France;
        {\texttt{\small name.surname@inrae.fr}}}%
}

\begin{document}

\maketitle
\thispagestyle{empty}
\pagestyle{empty}

\begin{abstract}
Energy management is a fundamental challenge for legged robots in outdoor environments. 
Endurance directly constrains mission success, while efficient resource use reduces ecological impact. 
This paper investigates how terrain slope and heading orientation influence the energetic cost of quadruped locomotion.
We introduce a simple energy model that relies solely on standard onboard sensors, avoids specialized instrumentation, and remains applicable in previously unexplored environments. 
The model is identified from field runs on a commercial quadruped and expressed as a compact function of slope angle and heading. 
Field validation on natural terrain shows near-linear trends of force-equivalent cost with slope angle, consistently higher lateral costs, and additive behavior across trajectory segments, supporting path-level energy prediction for planning-oriented evaluation. 
\end{abstract}


\section{INTRODUCTION}

Robots are increasingly deployed in demanding outdoor environments such as agriculture, mining, planetary exploration, and disaster response.  
In these domains, autonomy is limited not only by perception and decision-making capabilities but also by the ability to complete missions within available energy resources \citep{otsuEnergyAwareTerrainAnalysis2016}.
This challenge is particularly acute for mobile platforms operating across diverse terrains and conditions \citep{ahmadiRealtimeEnergyoptimalPath2024}.
Ensuring sufficient endurance is therefore a prerequisite for reliable and effective operation in the field.

Energy awareness is essential for both practical and ecological reasons.  
On the one hand, efficient energy use extends mission duration and reduces the risk of premature task interruption.  
On the other, sustainable practices in sectors such as agriculture require machines that minimize unnecessary energy expenditure and reduce their environmental footprint.  
Robots must therefore plan motions that simultaneously optimize operational efficiency and respect ecological constraints.  

Legged platforms offer unique advantages in natural and agricultural settings where wheeled or tracked vehicles may be less suitable. 
They can traverse irregular ground, adapt to varying terrain geometries, and operate in areas where heavy machines would damage vegetation or soil through compaction.  
These properties make quadrupeds and other legged systems attractive candidates for tasks that demand both mobility and low environmental impact \citep{bechar2016review,hutterANYmalaHighlyMobile2016}.  

\begin{figure}[t]
   \centering
   \includegraphics[width=0.48\textwidth]{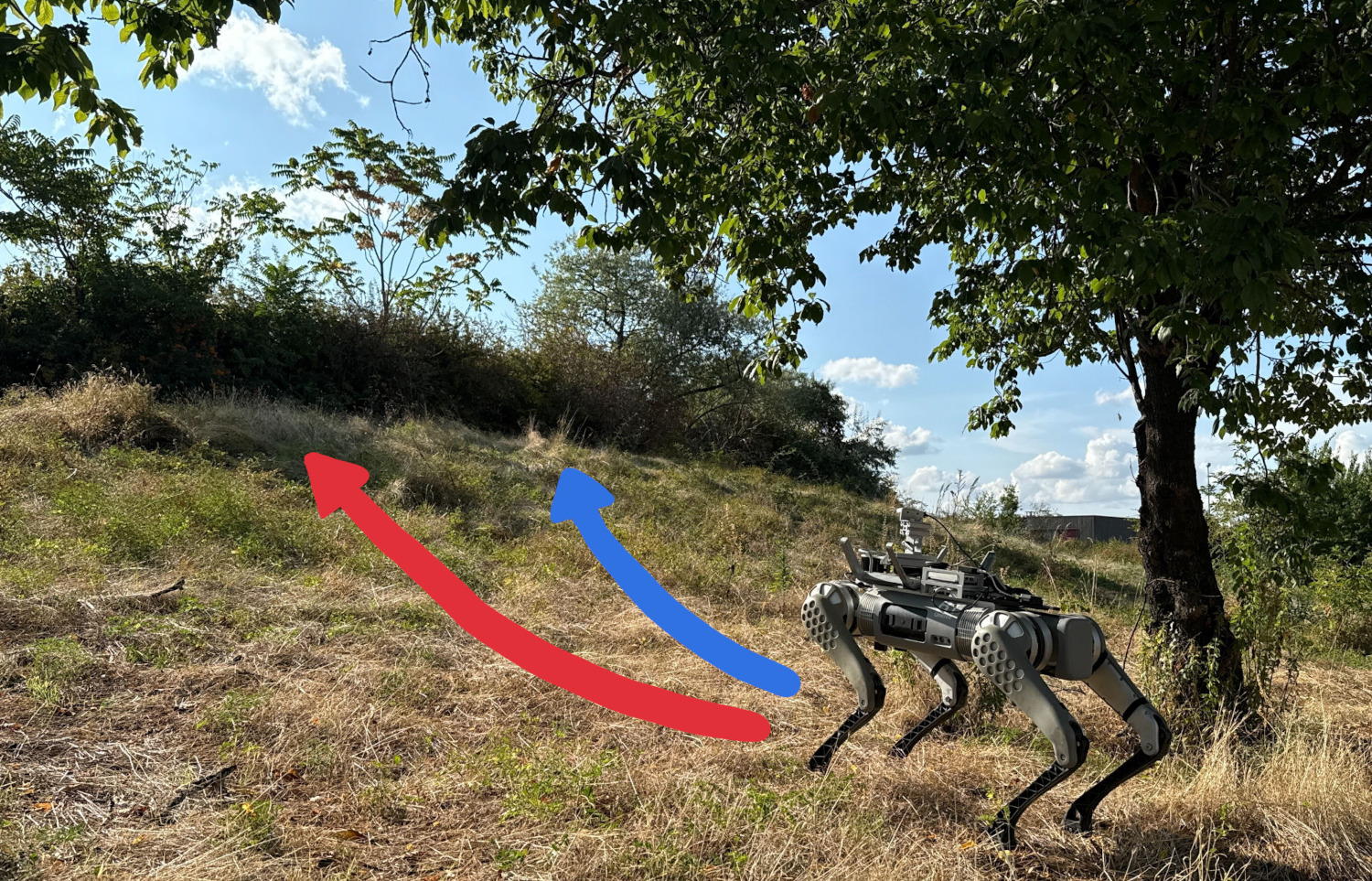}
   \caption{For legged robots, the complexity of their dynamics makes energy consumption difficult to predict. 
   The energy consumption of the two illustrated paths (blue: direct uphill, red: indirect detour) is not necessarily proportional to distance: on sloped terrain, longer but smoother trajectories can require less energy than shorter, steeper ones.
   }
   \label{fig:intro_scenario}
\end{figure}

However, as seen in \autoref{fig:intro_scenario}, the energy requirements of legged locomotion are difficult to predict.  
Unlike wheeled vehicles, where motion costs can often be related to distance and slope, legged robots rely on complex multi-body dynamics, coordinated gaits, and repeated ground contacts.  
This makes it challenging to identify which locomotion strategies are energetically most favorable under specific terrain conditions.  

This work focuses on the influence of terrain slope on the energy consumption of quadruped robots.  
We analyze how heading direction relative to an incline affects overall energy use and how such knowledge can inform motion planning.  
By modeling these relationships from onboard measurements, we introduce a practical basis for energy-aware navigation strategies in real-world agricultural and outdoor environments.  
To this end, our contributions are
\begin{itemize}
  \item A simple, generic, and reproducible method to predict heading-dependent energy maps for quadrupeds using only standard onboard sensors;
  \item A calibration procedure linking energy consumption to the robot's movements; and
  \item A field-validated model and an illustration of its relevance for energy-aware, planner-level path-cost evaluation on sloped terrain.
\end{itemize}

The remainder of this paper is organized as follows.  
\autoref{sec:related} reviews existing approaches to energy modeling, from wheeled to legged robots, and identifies the gap concerning heading-dependent costs in path planning.  
\autoref{sec:theory} introduces the proposed framework, including the notation, motion model, and the methods used to construct and learn energy maps as well as to evaluate candidate paths.  
\autoref{sec:experiments} presents field experiments that validate the modeling assumptions and illustrate the relevance of the model for planning-oriented path-cost evaluation.

\section{RELATED WORK}
\label{sec:related}

Energy-aware navigation has evolved along two largely separate paths: physics-based models for wheeled platforms and stability-focused approaches for legged systems. This split has created a critical gap in demanding outdoor environments where both energy efficiency and locomotion versatility are essential for mission success \citep{otsuEnergyAwareTerrainAnalysis2016}.

Early mobile robotics treated energy consumption as proportional to distance traveled, a reasonable approximation on flat terrain where rolling resistance dominates \citep{zhangSurveyEnergyefficientMotion2020}. However, this equivalence breaks down dramatically when robots encounter slopes. Gravitational work introduces fundamental asymmetries that depend on both terrain inclination and motion direction relative to the gradient \citep{otsuEnergyAwareTerrainAnalysis2016}. Classical energy models for wheeled vehicles decompose power into rolling resistance, aerodynamic drag, and gravitational components \citep{houEnergyModelingPower2018}. This immediately reveals that heading direction relative to slope matters as much as slope magnitude itself. The insight led to anisotropic cost fields where energy becomes a function of both position and heading direction \citep{sanchez-ibanezOptimalPathPlanning2023}. Planners can now exploit directional asymmetries for substantial energy savings. Such approaches demonstrate that even moderate slopes create opportunities for optimization through careful heading selection \citep{kivekasEffectSoilProperties2024}.

The fact that optimal paths depend on motion direction motivated sophisticated geometric approaches \citep{sanchez-ibanezOptimalPathPlanning2023}. These frameworks treat cost as an explicit function of both configuration and velocity direction \citep{bulloGeometricControlMechanical2004}. Fast marching methods and Hamilton-Jacobi formulations further enable efficient computation of direction-dependent optimal paths \citep{mirebeauHamiltonianFastMarching2019}. However, these mathematical tools remain largely disconnected from empirical energy measurements on legged platforms. This limits their applicability to real-world quadruped navigation \citep{roweObtainingOptimalMobilerobot1997a}.

Research on quadruped energetics has pursued a fundamentally different path, emphasizing gait optimization, mechanical efficiency, and cost of transport analysis primarily on level ground \citep{luneckasHexapodRobotGait2021}. Studies establish clear relationships between speed, coordinated gaits, and energy consumption for various legged platforms \citep{harperEnergyEfficientNavigation2019}. They provide insights into actuator efficiency and the role of leg compliance in reducing metabolic cost. Modern quadrupeds such as ANYmal have demonstrated robust rough-terrain locomotion with reasonable energetic cost \citep{hutterANYmalaHighlyMobile2016}. This pushes the analysis toward real-world conditions while maintaining focus on stability and traversability. However, energy considerations typically appear as learned penalties or uniform scalars applied to flat-ground costs rather than as primary variables that guide global path selection. Energy remains secondary to traversability \citep{mikiLearningRobustPerceptive2022}.

Recent work has nevertheless addressed energy efficiency in quadrupeds from different angles. Energy consumption can be reduced at the locomotion level by optimizing foot trajectories through reinforcement learning \citep{yan2024energy}. At the navigation level, proprioceptive signals, including current consumption, can also be used to estimate traversability for outdoor legged navigation \citep{elnoor2024pronav}. These contributions are related to ours, but differ in scope. Rather than learning a locomotion policy or predicting a generic traversability score, our objective is to derive a compact heading-dependent energy model on sloped terrain, intended for planner-level path-cost evaluation.

Recent advances in perceptive locomotion have enabled quadrupeds to navigate complex outdoor terrain with impressive robustness \citep{wellhausenRoughTerrainNavigation2021}. These systems use vision and proprioception to select footholds and avoid obstacles. They excel at local navigation and real-time adaptation to terrain features, yet global route choice remains driven by geometric heuristics rather than empirically grounded energy models \citep{mikiLearningRobustPerceptive2022}. Agricultural applications further highlight this gap, where robots must balance task completion with energy efficiency across varied terrain geometries \citep{oliveiraAdvancesAgricultureRobotics2021}. They lack the tools to predict how different paths will affect battery consumption.

Unlike wheeled vehicles that may recover energy through regenerative braking during descent, quadrupeds must actively control their limbs throughout the gait cycle \citep{harperEnergyEfficientNavigation2019}. This makes downhill motion energetically nontrivial. The fundamental difference suggests that energy-optimal paths for legged robots may involve complex heading strategies rather than simple elevation minimization. Evidence from planetary analogue experiments confirms that heading direction relative to slope can reverse energetic preferences \citep{otsuEnergyAwareTerrainAnalysis2016}. Indirect traverses can be competitive with direct climbs under certain conditions. 

Recent work has begun to address energy prediction for ground robots on uneven terrain, demonstrating that slope direction significantly affects consumption for wheeled platforms \citep{weiPredictingEnergyConsumption2022}. However, transfer to legged systems requires new measurements and validation because multi-body dynamics, coordinated gaits, and ground contacts fundamentally alter the energy budget compared to wheels. 

Many existing energy models assume access to joint torque sensors, detailed terrain geometry, or sophisticated mechanical models that complicate field deployment on commercial platforms \citep{herediaECDPEnergyConsumption2023}. While such signals provide rich information for research validation, practical autonomy requires approaches that leverage ubiquitous sensors such as battery management systems, odometry, and inertial measurement units \citep{liuOpenApproachEnergyEfficient2023}. Battery electrical power serves as a conservative proxy for mechanical work and directly relates to mission duration. This makes it attractive for energy-aware planning despite being less precise than torque-based estimates. As such, we propose a generic and easy-to-calibrate energy model that relies only on standard battery measurements.

\newcommand{\se}{\mathfrak{se}}
\newcommand{\bvarpi}{\bm{\varpi}}
\newcommand{\bforce}{\bm{f}}

\section{THEORY}
\label{sec:theory}

This section establishes the mathematical foundation for energy prediction on sloped terrain. We begin with notation and coordinate systems, describe the robot motion model, and present the theoretical tools for constructing energy models from empirical measurements.

\subsection{Notations and Assumptions}
\label{sec:notation}

\begin{figure}[htbp]
      \centering
      \includegraphics[width=\linewidth]{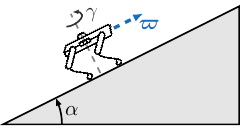}
      \caption{Coordinate system on a sloped terrain. The robot moves with velocity $\bvarpi(t)$ on a slope of angle $\alpha$, oriented with a heading $\gamma$ with respect to the slope direction.}
      \label{fig:slope_geometry}
   \end{figure}

Let $\alpha$ denote the local slope angle of the terrain, defined with respect to the horizontal plane. 
The robot's heading relative to the slope direction is denoted by $\gamma$, where $\gamma = 0^\circ$ corresponds to the robot facing directly uphill, $\gamma = 180^\circ$ to facing directly downhill, and $\gamma = 90^\circ$ to facing perpendicular to the slope. 
We assume that the robot evolves in a locally planar environment. As such, the generalized velocity (twist) of the robot is noted as $\bvarpi^\wedge\in\se(2)$, of which $\bvarpi\in\mathbb{R}^3$ is the coordinates of the twist \cite{barfoot2024state}.
\autoref{fig:slope_geometry} depicts a summary of the notations.

The coordinate system assumes that terrain can be locally approximated as planar over the length scales relevant to energy measurement. Real terrain exhibits complex three-dimensional structure, but this simplification captures the first-order effects that dominate energy consumption on moderate slopes. 

\subsection{Energy of a given path}
\label{sec:energy_given_path}

Having established the coordinate system and geometric relationships, we now formalize the energy computation for robot trajectories. We define the robot's path $\mathcal{P}$ as a function from time to the Special Euclidean group $SE(2)$ as
\begin{equation}
    \mathcal{P}: \mathbb{R} \rightarrow SE(2),
\end{equation}
where $\mathcal{P}(t) \in SE(2)$ represents the robot's pose (position and orientation) at time $t$. The associated body velocity $\bvarpi(t)^\wedge\in\se(2)$ is 
\begin{equation}
    \bvarpi(t)^\wedge = \mathcal{P}(t)^{-1} \frac{d}{dt} \mathcal{P}(t).
\label{eq:velocity}
\end{equation}

From this, the energy required to follow a given path from $t=0$ to $t=t_\text{goal}$ is given by the integral of the instantaneous power, as
\begin{equation}
    E = \int_0^{t_\mathrm{goal}} \left\langle \bforce(t)^\wedge \mid \bvarpi(t)^\wedge \right\rangle \, dt,
\end{equation}
where $\bforce(t)^\wedge \in \se^*(2)$ is the generalized force (wrench) applied at pose $\mathcal{P}(t)$, and $\bvarpi(t)^\wedge \in \se(2)$ is the generalized velocity.
The above equation simplifies to
\begin{equation}
    E = \int_0^{t_\mathrm{goal}} \bforce(t)^T \bvarpi(t) \, dt,
\label{eq:final_energy}
\end{equation}
which is the inner product between two real $3\times 1$ vectors integrated over time.

The overall difficulty comes from the complexity of the wrench $\bforce$, which may depend on many unknown and unobservable variables.
In the following, we assume that the wrench depends only on terrain slope and relative heading, so that


\begin{equation}
    \bforce : (\alpha,\gamma) \in \mathbb{R}^2 \mapsto \bforce(\alpha,\gamma) \in \se^*(2),
\end{equation}
where $\alpha \in [0, \pi/2)$ is the slope inclination and $\gamma \in [0, \pi]$ is the heading angle as defined previously. Note that $\gamma$ can be restricted to $[0, \pi]$ due to symmetry considerations (the energy cost of approaching a slope at an angle $\gamma$ is the same as approaching at an angle $2\pi - \gamma$).

Under this simplified model, a useful linearity property holds. If a velocity $\lambda\bvarpi_1$ (respectively $\bvarpi_2$) produces an instantaneous energy consumption of $P_1\,dt$ (respectively $P_2\,dt$), then the combined motion produces an instantaneous energy consumption of $(\lambda P_1 + P_2)\,dt$.

Although these assumptions are strong, they provide a practical approximation and make calibration more straightforward, since the problem reduces to fitting a low-dimensional function.

Under the terrain-and-heading assumption, the wrench depends on local slope angle and heading:
\begin{equation}
    \bforce(\alpha,\gamma)=\begin{bmatrix}f_x(\alpha,\gamma)\\[2pt] f_y(\alpha,\gamma)\\[2pt] \tau(\alpha,\gamma)\end{bmatrix}.
\label{wrench}
\end{equation}

At each instant, the body velocity is known and defined by~\autoref{eq:velocity}.  
The wrench components in~\autoref{wrench} are determined experimentally.  
With both quantities available, the path energy $E$ follows directly from~\autoref{eq:final_energy}.  
In summary, the energy prediction reduces to identifying the wrench components, 
thus closing the framework with a compact and practical formulation for path-level estimation.

\section{EXPERIMENTS}
\label{sec:experiments}

This section outlines an outdoor protocol to estimate, from onboard signals only, the components of the applied wrench and to assemble a dataset for future energy-aware navigation. 

\subsection{Experimental Setup}
\label{sec:setup}

We use a Unitree B1 with its stock sensor suite, shown in \autoref{fig:intro_scenario}: a battery monitor (voltage and current), an IMU that provides orientation, and legged odometry that yields pose and body-frame velocities. The experiments take place outdoors on a mix of a controlled ramp and natural grassy terrain. Slopes range roughly from \(5^\circ\) to \(20^\circ\).
The robot travels along straight segments at constant speed of \SI{0.3}{\m\per\s}, using its default walking gait. 

\subsection{Data Preprocessing}
\label{sec:acq-angles}
The data stream is continuous and minimal: IMU orientation, odometry with body-frame velocities, and battery voltage–current pairs. 
Very low-speed samples are discarded to avoid division artifacts. To limit noise and spikes, we apply a light median-type outlier filter and a short exponential moving average on both power and velocities. We also enforce a simple consistency bound between electrical and mechanical power to reject clearly inconsistent points.
The slope inclination \(\alpha\) and the heading direction \(\gamma\) are extracted from the IMU gravity vector. 

\subsection{Calibration and Evaluation Protocol}
\label{sec:protocol}

We follow the approach in \autoref{sec:theory}. Our goal is to recover the applied wrench from onboard signals only.
Over short time windows, we compute the electrical power from the battery readings and read body-frame velocities from the odometry. 
This gives the in-plane forces and the yaw torque without external sensors. 

To make the estimation robust, we run a grid of tests that covers both terrain and motion. Slopes range from mild to steeper values on a ramp and on natural grass. Headings include uphill, downhill, cross-slope in both directions, and several intermediate angles. Speed is kept close to a constant value along each segment, and we repeat each condition several times to check repeatability.

The outcome is a dataset of short-window wrench estimates indexed by slope and heading, together with simple quality checks (repeatability across runs, basic power consistency, and removal of obvious outliers). This dataset is the basis we use to describe how the wrench varies with terrain and direction.

\begin{figure}[htbp]
    \centering
    \includegraphics[width=\linewidth]{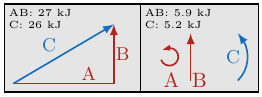}
  \caption{Energy superposition: measured energy of composite paths compared with the sum of their parts. Close agreement supports the additivity assumption used by the model.}
    \label{fig:energy_superposition}
\end{figure}


\begin{figure*}[!t]
  \centering
  \includegraphics[width=\linewidth]{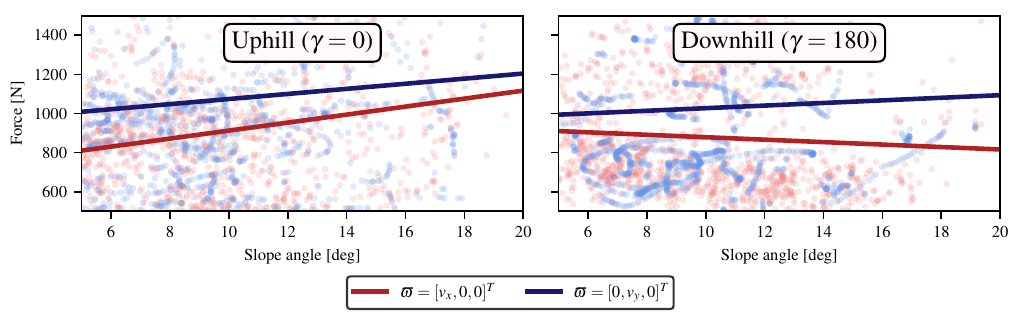}
  \caption{Measured force-equivalent cost as a function of slope angle \(\alpha\) for uphill (left) and downhill (right) runs. Red and blue points correspond to two velocity headings, \(\bvarpi(t) = [1,0,0]^T\) (forward) and \(\bvarpi(t) = [0,1,0]^T\) (lateral). The uphill panel shows increasing cost with slope angle, while the downhill panel shows a decrease only for forward motion. By contrast, lateral motion remains more expensive and tends to worsen with slope. These opposite trends indicate that energetic cost depends not only on terrain inclination but also on robot orientation with respect to the slope.}
  \label{fig:map_filter}
\end{figure*}

\subsection{Superposition and Path Equivalences}
\label{sec:superposition}

The model treats energy as the line integral of a local per-distance cost; therefore, energies add when path segments are concatenated. We test this superposition in two settings, both shown in \autoref{fig:energy_superposition}. On the left, the robot executes a two-segment path (A then B) and we compare its total to a single straight segment C that connects the same endpoints; the relative difference is about \(4\%\). On the right, we produce the same pose change either by “in-place yaw then straight” (A then B) or by a smooth circular arc C; again the relative difference is about \(13\%\). Across repeats, the differences stay within the natural variability of the measurements. These results are consistent with the per-distance formulation and support linear accumulation of energy along simple motion primitives.

\subsection{Force-equivalent cost versus slope angle}
\label{sec:power-slope}

As intuition would tell, the uphill panel of \autoref{fig:map_filter} shows the force-equivalent cost increasing with slope angle \(\alpha\), as the gravity component grows. Downhill, we observe the opposite trend, but only for forward motion. Lateral motion remains costly and tends to worsen with slope, which is consistent with the additional effort required to preserve stability during cross-slope progression. In both downhill cases, despite the favorable gravitational component, a significant fraction of the mechanical energy is dissipated through non-conservative effects, notably impacts, structural deformations, and heat losses arising from repeated ground contacts.

For planning, two practical observations follow directly. First, ascent and descent have opposite trends, reflecting the sign change in the gravity contribution. Second, the gap between red and blue curves shows that heading is as important as slope magnitude. The cost of motion is therefore jointly determined by slope angle and heading. Going downhill is not energetically free, even though it remains easier than climbing when the robot stays aligned with the slope.

\begin{figure}[htbp]
  \centering
  \includegraphics[width=\linewidth]{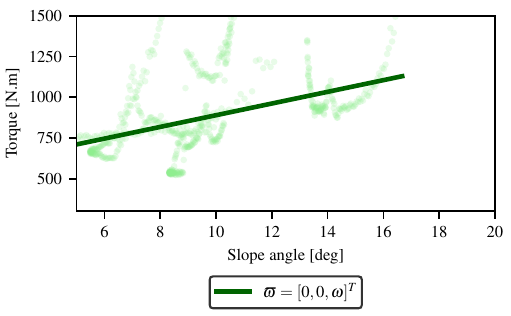}
  \caption{Estimated yaw torque as a function of slope angle \(\alpha\) for pure rotation movements. The trend suggests an increase of rotational cost with slope.}
  \label{fig:energy_rotation}
\end{figure}

The same reasoning can be extended to pure rotations, as shown in \autoref{fig:energy_rotation}. Given the instantaneous power \(P\) and rotational velocity \(\omega\), the yaw torque estimate is \(\tau = P/\omega\). The overall trend indicates that rotational cost increases with slope angle. This result should be read as preliminary, but it is consistent with the idea that reorienting the robot on an incline requires additional stabilization effort.

While the present results provide a simple model for describing the energy expenditure of a quadruped robot, the fit with experimental data remains limited, indicating that several parameters are not yet captured. Nevertheless, such a simplified model may still prove sufficient for high-level planning purposes.
Additional experiments are still required to validate the model and assess its generality across operating conditions. 
\section{CONCLUSION}

We presented a simple model for heading-dependent energy prediction on sloped terrain and showed that it can be identified from standard onboard measurements. Field experiments validate three main findings: a clear dependence of force-equivalent cost on slope angle and heading, additive behavior across composed trajectory segments, and a preliminary increase of rotational cost with slope. Together, these results support planning-oriented, path-level energy evaluation for legged robots in outdoor conditions.

Future work will extend this framework by coupling online slope estimation with path-cost evaluation, and by broadening identification across velocities, gaits, terrains, and robot platforms.

\section*{ACKNOWLEDGMENT}


This work was funded by the French National Research Agency (ANR) through the Junior Research Chair program. It also received support from the International Research Center “Innovation Transportation and Production Systems” of the I-SITE CAP 20-25, under a scholarship grant.

\section*{REFERENCES}

\renewcommand*{\bibfont}{\small}
\printbibliography[heading=none]



\end{document}